\documentclass[letterpaper]{article} 
\usepackage{aaai25}  
\usepackage{times}  
\usepackage{helvet}  
\usepackage{courier}  
\usepackage[hyphens]{url}  
\usepackage{graphicx} 
\usepackage{natbib}  
\usepackage{caption} 
\usepackage{algorithm}
\usepackage{algorithmic}
\usepackage[algo2e]{algorithm2e} 
\usepackage{amsmath}
\usepackage{newfloat}
\usepackage{listings}
\DeclareCaptionStyle{ruled}{labelfont=normalfont,labelsep=colon,strut=off} 
\floatstyle{ruled}
\newfloat{listing}{tb}{lst}{}
\floatname{listing}{Listing}
\usepackage{graphicx}
\usepackage{booktabs}
\usepackage{times}
\usepackage{multirow}
\usepackage{epsfig}
\usepackage{amssymb}
\usepackage{booktabs}
\usepackage{verbatim}

\usepackage{enumitem}
\usepackage{pifont}
\usepackage{listings}%
\usepackage{bbding}
\usepackage[accsupp]{axessibility}
\newcommand{\bigO}{\mathcal{O}}
\title{FreeCap: Hybrid Calibration-Free Motion Capture in Open Environments}

\author{
    Aoru Xue\textsuperscript{\rm 1,}\equalcontrib,
    Yiming Ren\textsuperscript{\rm 1,}\equalcontrib,
    Zining Song\textsuperscript{\rm 1},
    Mao Ye\textsuperscript{\rm 2},
    Xinge Zhu\textsuperscript{\rm 3},
    Yuexin Ma\textsuperscript{\rm 1,}\thanks{Corresponding author.}
}
\affiliations{
    \textsuperscript{\rm 1}ShanghaiTech University\\
    \textsuperscript{\rm 2}Inceptio Technology\\
    \textsuperscript{\rm 3}Shanghai Jiao Tong University\\
    \{xuear2024,~renym2022,~mayuexin\}@shanghaitech.edu.cn
}

\usepackage{bibentry}

\begin{document}
\makeatletter
\let\@oldmaketitle\@maketitle
\renewcommand{\@maketitle}{
   \@oldmaketitle
  }
\maketitle
\begin{abstract}
We propose a novel hybrid calibration-free method \textbf{FreeCap} to accurately capture global multi-person motions in open environments. Our system combines a single LiDAR with expandable moving cameras, allowing for flexible and precise motion estimation in a unified world coordinate. In particular, We introduce a local-to-global pose-aware cross-sensor human-matching module that predicts the alignment among each sensor, even in the absence of calibration. Additionally, our coarse-to-fine sensor-expandable pose optimizer further optimizes the 3D human key points and the alignments, it is also capable of incorporating additional cameras to enhance accuracy. Extensive experiments on Human-M3 and FreeMotion datasets demonstrate that our method significantly outperforms state-of-the-art single-modal methods, offering an expandable and efficient solution for multi-person motion capture across various applications.
\end{abstract}
\begin{figure*}[t!]
	\centering
	\includegraphics[width=\linewidth]{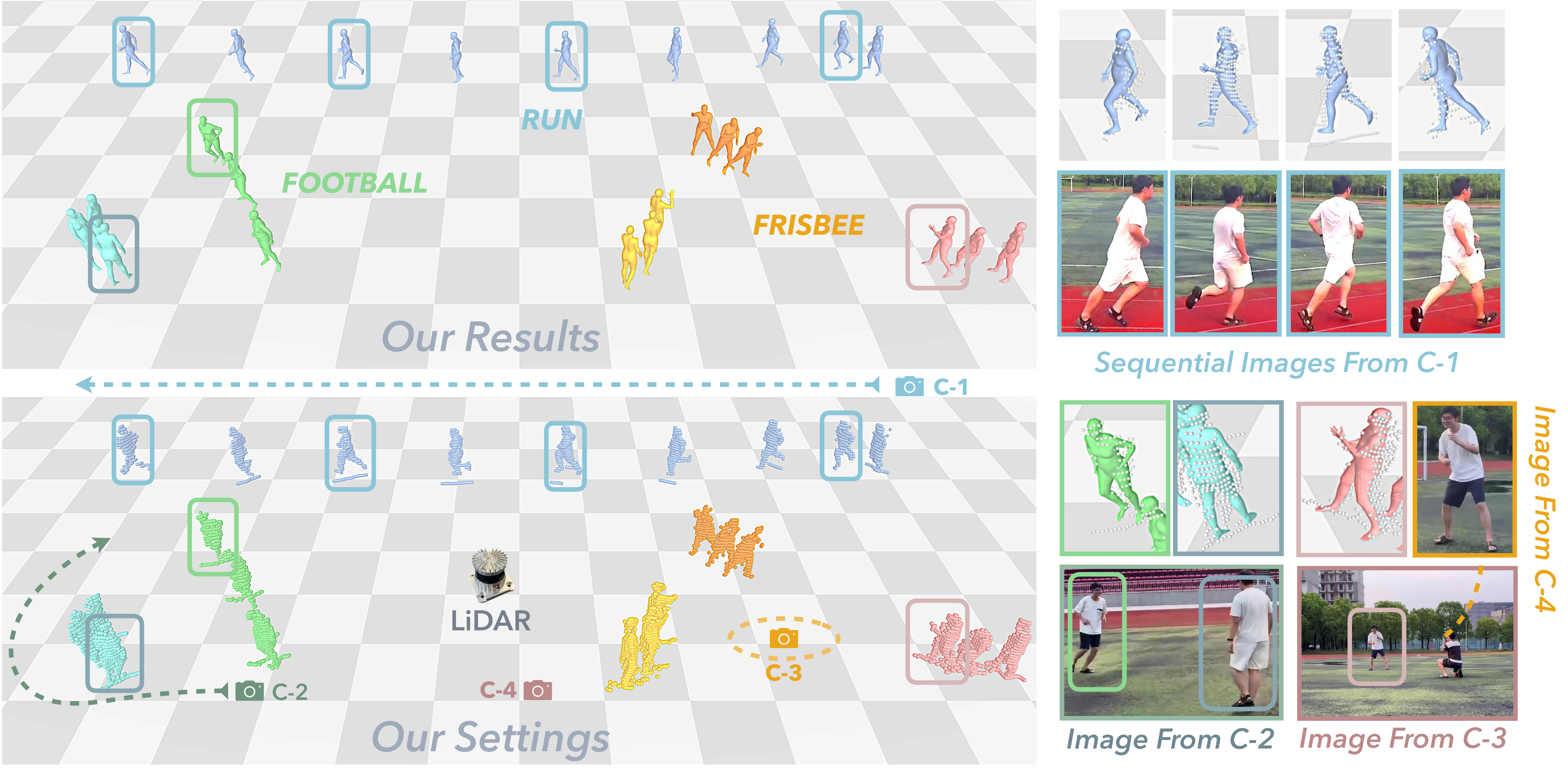}
	\caption{Visualization of our FreeCap in a real-time captured scenario. Our settings include a single LiDAR and four cameras. Camera-1 follows the running person, camera-2 surrounds two people playing soccer, camera-3 focuses on the main person playing frisbee and camera-4 captures three persons. We zoom in some cases to the right.}
	\label{fig:teaser}
\end{figure*}
\section{Introduction}
Vision-based human motion capture refers to the process of accurately predicting and reconstructing human meshes in open environments by utilizing visual data from LiDAR or camera. The approach eliminates the need for wearable sensors, enabling the capture of natural and unrestricted human motion. It is particularly advantageous in scenarios such as daily activities and sports, where the use of wearable devices may be impractical or restrictive.

Camera-based methods~\cite{BelagZ2016, Miezal2017, rajasegaran2022tracking, wham:cvpr:2024} can effectively capture the accurate human local pose by image texture information but lack depth information, and depth camera-based methods~\cite{SridhMOT2015, Wei:2012, BodyFusion, DoubleFusion, HybridFusion} suffer from lighting conditions and the limited sensor range. 
LiDAR is widely used in robotics and autonomous driving~\cite{zhu2021cylindrical,zhu2020ssn,cong2022stcrowd,xu2023human,lu2023see} for perceiving depth information in large-scale scenes. 
Previous LiDAR-based methods~\cite{li2022lidarcap, fan2023lidar,ren2024livehps} can estimate the global human pose in large-scale scenes by accurate depth information. However, the sparsity of long-distance point clouds and the lack of texture information can lead to inaccuracies in predictions.

To leverage both image texture and LiDAR depth information for more accurate global human motion predictions in open environments. The previous method\cite{Cong_Xu_Ren_Zhang_Xu_Wang_Yu_Ma_2023,fan2023human} utilizes calibrated multiple cameras and LiDARs for involving the fusion of raw data, which can lead to difficulties in unifying data distribution and reduced generalization ability. In contrast, as shown in Fig.~\ref{fig:teaser}, our system consists of the single LiDAR and any number of moving cameras. The LiDAR captures extensive global depth information, while the moving cameras provide detailed texture information in local areas. This more flexible and expandable setup enables accurate multi-person motion estimation within a unified world coordinate system, making it suitable for diverse applications.

In this paper, we propose a novel calibration-free multi-model method \textbf{FreeCap}. We can not directly get the cross-sensor human matching in large-scale multi-person scenes without the calibration and the data-driven method can not intentionally learn the calibration when the camera is moving. To address these challenges, we introduce an efficient and high-quality local-to-global matching module, Pose-aware Cross-sensor Matching. Specifically, we use RTMPose~\cite{jiang2023rtmpose} and WHAM~\cite{wham:cvpr:2024} to estimate the 2D key points and body pose in image and use LiveHPS~\cite{ren2024livehps} to estimate the 3D key points and body pose in point cloud. The body poses are used for local matching and the global key points are used for matching refinement. Next, we design a coarse-to-fine sensor-expandable pose optimizer to optimize the coarse align matrix calculated based on the matched 2D key points and 3D key points, which can unify the calibration-unknown multi-modal data and guide the network to further optimize the 3D human key points. Furthermore, considering the narrow of the camera range, our network can accept expandable camera data, the more cameras can assist LiDAR to predict more accurate results. Extensive experiments conducted on the multi-person large-scale dataset Human-M3\cite{fan2023human} and the multi-view sensor dataset FreeMotion\cite{ren2024livehps}, demonstrate that our method achieves significant improvements in human pose compared to other single-modal SOTA methods.

Our main contributions can be summarized as follows:
\begin{itemize}
    \item[$\bullet$] We present the first calibration-free and sensor-expandable system designed for capturing multi-person motions in open environments.
    \item[$\bullet$] We propose a local-to-global cross-sensor human-matching approach for predicting alignment matrix.
    \item[$\bullet$]We design an effective multi-modal coarse-to-fine fusion method by optimizing human key points.
    \item[$\bullet$] Our method achieves SOTA on the large-scale dataset Human-M3 and multi-sensor dataset FreeMotion.
\end{itemize}

\section{Related Work}
\subsection{Wearable Sensor-based Methods}
Early motion capture system reconstructs human mesh relays on dense markers ~\cite{loper2014mosh,Park2008,SongGodoy2016,raskar2007prakash, UnstructureLan} attach in human body. It enables the capture of high-quality human motions and is wildly applied in industry. However, system is costly and limited in indoor scenes with good lighting conditions. The freer system uses inertial measurement units~\cite{yi2021transpose,PIPCVPR2022,ren2023lidar} for capturing orientations and accelerations of human key bones which reflect the human motion representation based on SMPL model. It can work in relatively open scenarios, but it suffers from drift and can be affected by magnetic field. Moreover, above methods require actors to wear sensors, causing constrained human motions.

\subsection{Camera-based Methods}
To make mocap method applied in daily usage, the markerless methods~\cite{liu2013markerless, Rhodin:2016, hmrKanazawa17} use cameras only has made great progress. The multi-view camera-based methods~\cite{malleson2019real,zhang2020fusing,malleson2020real} predict accurate human motions based on optimization from full perspective motion information. Though performers do not need to wear markers, the multi-camera studio needs complex deployment in closed areas. The monocular-based method~\cite{mehta2017monocular,Sapp2013} only requires a single camera, it is light-weight while the performer is free to move, but a single camera can not perceive depth information and the sensor range is narrow. Recently, some methods~\cite{wham:cvpr:2024, rajasegaran2022tracking}, begin to pay attention to estimating the global human trajectory from a monocular dynamic camera. However, the trajectory is relative to a beginning frame and accumulated errors occur, which means that methods cannot infer long-term data and cannot predict multi-person global motions in a unified world coordinate system.
\begin{figure*}[t!]
	\centering
	\includegraphics[width=\linewidth]{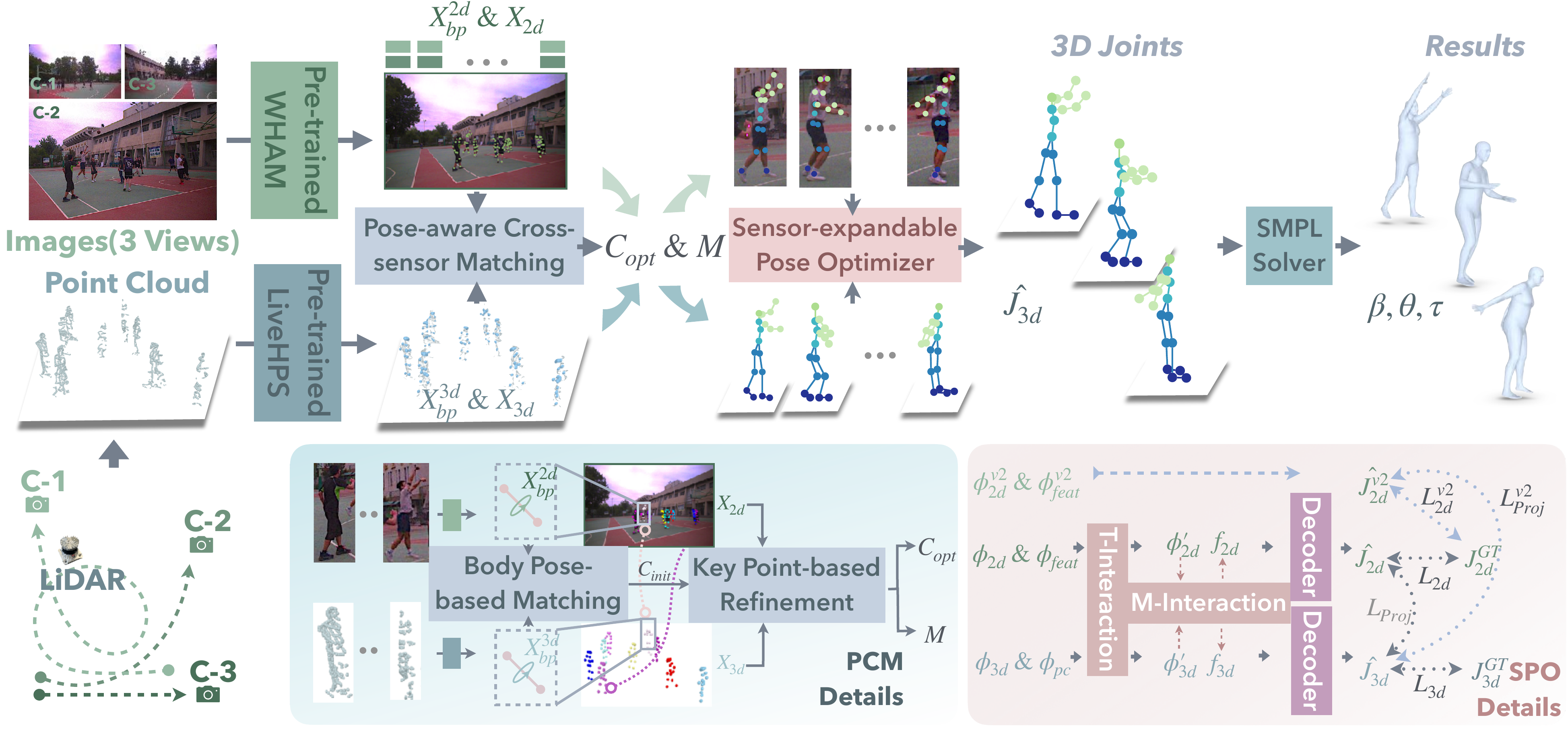}
	\caption{The pipeline of FreeCap. It consists of three main modules, including the pose-aware cross-sensor matching estimate of the optimal pairs and aligns matrix, the sensor-expandable pose optimizer predicts the 3D human joints, and the SMPL solver regresses the SMPL parameters. We also show the details of PCM and SPO.}
	\label{fig:pipeline}
\end{figure*}
\subsection{LiDAR-based Methods}
In order to estimate the global human motions in large-scale open environments, various LiDAR-based mocap methods have been proposed. LiDARCap~\cite{li2022lidarcap} first proposes a single LiDAR-based human motion capture method, but it only estimates the local pose based on the graph-based framework. LiDAR-HMR~\cite{fan2023lidar} utilizes the point cloud geometric information to reconstruct the human mesh, but it can not provide the template-based motion representations, the scope of application is limited. PointHPS~\cite{cai2023pointhps} provides a cascaded network to predict the human pose and shape, but the network framework requires dense points in close range. LiveHPS and LiveHPS++~\cite{ren2024livehps,ren2025livehps++} estimate the full SMPL parameters, including the human local pose, body shape, and global translation, fully utilizing the depth information of point cloud. However, the sparsity of data and the lack of texture information also lead to inaccurate predictions.

\subsection{Hybrid Sensor-based Methods}
As all single modality sensor-based methods have corresponding disadvantages, recent works begin to explore the hybrid sensor-based method. Some methods utilize cameras and IMUs~\cite{liang2022hybridcap} to estimate more accurate human local poses, but still suffer from loss of depth information. LIP~\cite{ren2023lidar} uses the single LiDAR to provide depth information and four IMUs attached in for limbs to optimize the occlusion case. Human-M3~\cite{fan2023human} proposes a camera-LiDAR fusion method, but it relies on accurate calibrations causing the system must be fixed and pre-deployment. We present the first calibration-free method based on the single LiDAR and expandable cameras, which can work in open environments.

\section{Methodology}
Our goal is to capture human motion based on a single LiDAR with expandable moving cameras in the open environment. An overview of our pipeline is shown in Fig.~\ref{fig:pipeline}. We take the sequential point clouds and multi-view videos as input. There are three main modules in our network, including pose-aware cross-sensor matching(PCM), sensor-expandable pose optimizer(SPO), and the SMPL solver. Firstly, we use PCM to get the matching pairs between point cloud and the images. Then, based on the matching pairs and predicted human key points, we use an optimization framework to generate the alignment matrix, which can unify the cross-sensor data into the unified world coordinate. Next, the SPO takes expandable unified data as the input and optimizes the 3D key points by multi-sensor fusion. Finally, we use the SMPL solver to estimate the SMPL parameters from refined 3D key points.

\subsection{Preliminaries}
Our framework takes sequential multi-person’s point clouds and videos as input and predicts all person’s SMPL parameters. For PCM, we define $x_{3d}(n, t)$ and $x_{2d}(n, t)$ to represent 3D and 2D key points; $x_{bp}^{3d}(n, t)$ and $x_{bp}^{2d}(n, t)$ as body poses extracted by pretrained model; $C(n, t)$ as matching pairs between LiDAR and camera for $n\in N$ individuals in the scene. For SPO, we define $x_{pc}$ as normalized 3D point clouds; $x_{3d}$ as 3D key points which is root-centered; $x_{2d}$ as normalized 2D key point. The predicted 3D joints $\hat{J}_{3d}$ and 2D key points $\hat{J}_{2d}$ are supervised by the ground truth $J_{3d}^{GT}$ and $J_{2d}^{GT}$ respectively. We normalize $J_{3d}^{GT}$ and $J_{2d}^{GT}$ using the same normalization process applied to the input data following LiveHPS and WHAM. For the motion, We define $\theta^{GT}(n, t)$, $\beta^{GT}(n)$, and $T^{GT}(n, t)$ as the ground truth SMPL parameters, $N_J=24$ and $N_V=6890$ represents the number of human joint and mesh vertex.

\subsection{Pose-aware Cross-sensor Matching}
\label{subsec:PCM}
Human matching among multi-sensors is a prerequisite for our calibration-free method. We define the matching target is to get the matching pairs $C(t) = \{(i, j), i \in N, j \in M\}$,
where $N$ and $M$ are the sets of indices for the human appears in the LiDAR and camera respectively. And we use $\mathbf{C}_I$ to represent identity matching. The body pose, inherently independent of coordinate systems, is ideal for cross-sensor matching applications. Yet, relying solely on body pose might not capture global information accurately. For instance, when individuals perform synchronized activities, body pose alone may not provide precise matches. To address this, we introduce \textbf{Pose-aware Cross-sensor Matching(PCM)}, which integrates both global key point and local body pose data, as outlined in Alg.\ref{alg:algorithm}.

\begin{algorithm}[H]
\caption{Pose-aware Cross-sensor Matching}
\KwIn{$\{\mathbf{x}_{bp}^{3d}, \mathbf{x}_{bp}^{2d}, \mathbf{x}_{3d}, \mathbf{x}_{2d}, K, n_{iter}\}$}
\KwOut{$\mathbf{C}$}

$\mathbf{C}_{init}, \mathbf{M}_{init} \gets \text{Hungarian}(\text{Sim}(x_{bp}^{3d}, x_{bp}^{2d}))$\;

$n_{3d} \gets \text{len}(\mathbf{x}_{3d})$;
$n_{2d} \gets \text{len}(\mathbf{x}_{2d})$;

$v \gets \text{Variance}(\mathbf{M}_{init}^{transl})$\;

\If{$v > \delta $}{
    $\mathbf{Q} \gets \text{zeros}(n_{3d}, n_{2d})$\;

    \For{$t \gets 1$ \textbf{to} $T$}{
   
        $\mathbf{C}_{opt}, M, cost \gets \text{OptMatch}(\mathbf{x}_{3d}(t), \mathbf{x}_{2d}(t), \mathbf{x}_{bp}^{3d}(t),  \\ \mathbf{x}_{bp}^{2d}(t), K_t, n_{iter})$\;

        \ForEach{$(i, j) \text{ in } \mathbf{C}_{opt}$}{
            $\mathbf{Q}[i, j] \gets \mathbf{Q}[i, j]+ \text{cost} $\;
        }
        
    }

    $\mathbf{C} \gets \text{Hungarian}(\mathbf{Q})$\;
    
    \Return $\mathbf{C}$\;
}
\Else{
    \Return $\mathbf{C}_{init}$\;
}
\label{alg:algorithm}
\end{algorithm}

\subsubsection{Body Pose-based Matching}

We use pre-trained LiveHPS to predict 3D key points $x_{3d}(n, t)$ and human body pose $x_{bp}^{3d}(n, t)$, RTMPose to predict 2D key points $x_{2d}(n, t)$ and WHAM to predict human body pose $x_{bp}^{2d}(n, t)$, which are then employed to predict the matching pairs $C(t) = \{(i, j), i \in N, j \in M\}$ between LiDAR and cameras.
Then, we use sequential body poses $\mathbf{x}_{bp}^{3d}(n, t)$ and $\mathbf{x}_{bp}^{2d}(n, t)$ to calculate the body pose similarity for cost so that we can set up the Hungarian algorithm's cost matrix $\mathbf{Q}$, identifying a preliminary match $C_{init}$ and a corresponding calibration matrix $M_{init}$. The similarity $\text{Sim}(x_{bp}^{3d}(n), x_{bp}^{2d}(m))$ between the 3D body pose of person index $n$ and the 2D body pose of person index $m$ is calculated as follows:
\begin{equation}
    \begin{aligned}
        \text{sim}(x_{bp}^{3d}(n), x_{bp}^{2d}(m)) = \frac{1}{T} \sum_{t=1}^T \frac{x_{bp}^{3d}(n, t) \cdot x_{bp}^{2d}(m, t)}{\|x_{bp}^{3d}(n, t)\| \|x_{bp}^{2d}(m, t)\|}.
    \end{aligned}
\end{equation}

The matching cost $\mathbf{Q}$ is updated across all frames to secure the globally optimal matches using the Hungarian algorithm. Due to occlusions or sensor limitations, not all detected persons are matched; those with large re-projection errors are excluded to ensure optimal pairing. Additionally, to correct calibration errors potentially caused by the symmetric nature of human anatomy and to ensure the calibration reflects true orientations, we integrate global pose data. This approach also involves verifying matches by analyzing the variance $v$ of the calibration's estimated translation over multiple frames. High variance re-projection errors are smoothed using a temporal window. 

\subsubsection{Key Point-based Optimization Matching}

To further exploit the information of points in the scene and achieve more accurate global matching results, we propose Key-Point based Optimization Matching, enhance the matching and calibration accuracy using 2D and 3D key points $\mathbf{x}_{2d}$ and $\mathbf{x}_{3d}$

Given the candidate match $C \in \mathbf{C}$, we can use the Perspective-n-Point (\text{PnP}) algorithm to estimate the camera pose $M_C = \text{PnP}(x_{3d}, x_{2d}, C)$ that best aligns the paired 2D key points in the image with their corresponding 3D key points in world coordinates. Based on this estimated camera pose and the camera intrinsic matrix $K$, we define a projection matrix:
\begin{equation}
  P_C = K*M_C,
\end{equation}
which captures the transformation from 3D world coordinates to 2D image coordinates. 

Given the initial calibration matrix $M_C$, we define a function $E_{proj}(x_{3d}, x_{2d}, C)$ that calculates the re-projection error associated with that matrix:
\begin{equation}
E_{\text{proj}}(x_{3d}, x_{2d}, C) = \parallel P_C x_{3d}(C) - x_{2d}(C)\parallel_2.
\end{equation}
Thus, the problem of finding the optimal match can be formulated as:
\begin{equation}
C^* = argmin_{C \in \mathbf{C}}\{ E_{proj}(x_{3d},x_{2d},C)\}.
\end{equation}
Based on this, we can iteratively optimize the matching and calibration matrix according to the initial calibration matrix or initial matching. For the optimization process, we initialize proposal matches $\mathbf{C}_{proposal}$ with every 2D-3D pair first and optimize the proposal matches iteratively. We selected the set with the minimum weighted re-projection error and body pose error as the globally optimal matching. The weighted re-projection error and body pose error are defined below:
\begin{equation}
    \begin{aligned}
    E_{proj}
    (x_{3d},x_{2d}, x_{bp}^{3d}, x_{bp}^{2d}, C) &=E_{proj}(x_{3d}, x_{2d}, C) \\
    &+ \lambda_0 E_{proj}^{bp}(x_{bp}^{3d}, x_{bp}^{2d}, C), 
    \end{aligned}
\end{equation}
where the body pose matching error $E_{proj}^{bp}(x_{bp}^{3d}, x_{bp}^{2d}, C)$ is defined as below:
\begin{equation}
    \begin{aligned}
E_{proj}^{2d}(x_{bp}^{3d}, x_{bp}^{2d}, C) &=  \parallel J_{proj}(x_{bp}^{3d}, C) - J_{proj}(x_{bp}^{2d}, C)\parallel_2,
    \end{aligned}
\end{equation}
where $J_{proj}(x_{bp}^{3d}, C)$ and $J_{proj}(x_{bp}^{2d}, C)$ are the joint of SMPL with mean shape and body pose $x_{bp}^{3d}$, we project it to image plane with camera intrinsic matrix $K$. 
The detailed optimization process is presented in Alg.~\ref{alg:opt_match}. 

Our proposed Body Pose-aware Global Matching method leverages the distinctive qualities of body poses to bridge the sensorial gaps in multi-sensor environments. This method not only utilizes local body pose data but also incorporates global key point information, providing a robust framework for achieving accurate cross-sensor matching. By integrating both local and global data points, our approach significantly enhances the precision and reliability of the matching process, even in complex scenarios where traditional methods based solely on local data might fail.

\begin{algorithm}[H]
\caption{OptMatch}
\KwIn{$\mathbf{x}_{bp}^{3d}$, $\mathbf{x}_{bp}^{2d}$, $\mathbf{x}_{3d}$, $\mathbf{x}_{2d}$, $\mathbf{K}$, $n_{iter}$}
\KwOut{$\mathbf{C}_{opt}$, $M$, $\text{cost}_{max}$}

$n_{3d} \gets \text{len}(\mathbf{x}_{3d})$;
$n_{2d} \gets \text{len}(\mathbf{x}_{2d})$;

$\mathbf{C}_{\text{proposal}} \gets \emptyset$;
$\text{cost}_{max} \gets 0$;
$\mathbf{C}_{opt} \gets \emptyset$;

\For{$i \gets 1$ \textbf{to} $n_{3d}$}{  
    \For{$j \gets 1$ \textbf{to} $n_{2d}$}{  
        $\mathbf{Q} \gets -E_{proj}(\mathbf{x}_{3d}[i], \mathbf{x}_{2d}[j], \mathbf{C}_I) $\;  
          
        $\mathbf{C} \gets \text{Hungarian}(\mathbf{Q})$\;  
          
        $\mathbf{C}_{proposal} \gets \mathbf{C}_{proposal} \cup \{\mathbf{C}\}$\;  
          
    }  
} 

\ForEach{$\mathbf{C}$ \textbf{in} $\mathbf{C}_{proposal}$}{
    \For{$i_{iter} \gets 1$ \textbf{to} $n_{iter}$}{
        $ \mathbf{Q} \gets - E_{proj}(\mathbf{x}_{3d}, \mathbf{x}_{2d}, \mathbf{x}_{bp}^{3d}, \mathbf{x}_{bp}^{2d}, \mathbf{C}) $
        
        $ \text{cost} \gets \text{Sum}(\mathbf{Q}) $

        \If{$\text{cost} > \text{cost}_{max}$}{
            $\text{cost}_{max} \gets cost$ ;
            $\mathbf{C}_{opt} \gets \mathbf{C}$\;
        }
        
        $\mathbf{C} \gets \text{Hungarian}(\mathbf{Q})$\;
    }
}

$ M \gets \text{solvePnP}(\mathbf{x}_{3d}, \mathbf{x}_{2d}, \mathbf{C}_{opt})$

\Return $\mathbf{C}_{opt}$, $M$, $\text{cost}_{max}$\;
\label{alg:opt_match}
\end{algorithm}
\subsection{Sensor-expandable Pose Optimizer}
\label{subsec:SPO}
Each individual's 3D and 2D data are matched and used to compute a preliminary calibration matrix. We avoid network overfitting from implicit multi-modal calibration by transforming the data from the LiDAR coordinate system to the camera coordinate system using this matrix. Then we introduce the \textbf{Sensor-expandable Pose Optimizer(SPO)} for optimizing the human 3D joints by effectively integrating data from each sensor.

The SPO's inputs are processed by three specialized MLP encoders: a 3D motion encoder $E_{3d}$, a 2D motion encoder $E_{2d}$, and a 3D point cloud encoder $E_{pc}$. These encoders respectively transform 3D key points $x_{3d}$, 2D key points $x_{2d}$, and 3D point clouds $x_{pc}$ into feature representations $\phi_{3d}$, $\phi_{2d}$, and $\phi_{pc}$.

\subsubsection{Feature Integrator}
Lifting from 2D key points to 3D is inherently ambiguous. Therefore, we also utilize the static image features $\phi_{feat}$ extracted by a model, pre-trained on human mesh recovery task, as input, and integrate them with 2D feature integrator network $I_{2d}$ to enhance the 2D motion features $\phi_{2d}'$. 
The feature integrator employs residual connection layer with which the network can more easily learn the complex relationships between these features, resulting in a more expressive and comprehensive feature representation. Following WHAM, we pre-train the 2D motion encoder in SURREAL dataset with arbitrary camera motion. Similarly, point clouds and 3D key points features are integrated in the same way and get the enhanced feature $\phi_{3d}'$.
\subsubsection{Cross Modal Interaction}
Considering human motions are coherent over time, we employ temporal interactions to leverage the sequential nature of the data to capture temporal dependencies and correlations, thereby improving the overall results. We use multi-head self-attention layer to capture this temporal relationship and get the encoded feature $f_{3d}$ and $f_{2d}$. 
Subsequently, we interact the features from the arbitrary two modalities(or two sensor features) by designing a bidirectional cross-attention module and get the $f_{3d}$ and $f_{2d}$ from two distinct modalities.

Finally, we input all the 3D and 2D features into the Motion Decoder $D_{3d}$ and $D_{2d}$, where it predicts the 3D key points $\hat{J_{3d}}$ and 2D key points $\hat{J_{2d}}$ in their respective camera coordinate systems. To refine the results of 3D and 2D key points based on their projection correspondence, we add a regularization term for the re-projection loss.
The loss functions of our method are defined below:
\begin{equation}
    \begin{aligned}
    \mathcal{L}_{3d} &= \parallel \hat{J_{3d}} - J_{3d}^{GT}\parallel_2^2,\\
    \mathcal{L}_{2d} &= \parallel \hat{J_{2d}} - J_{2d}^{GT}\parallel_2^2,\\
    \mathcal{L}_{proj} &= \parallel Porj(\hat{J_{3d}}, K) - J_{2d}^{GT}\parallel_2^2,\\
    \mathcal{L}_{total} &= \lambda_1 \mathcal{L}_{3d} + \lambda_2 \mathcal{L}_{2d} + \lambda_3 \mathcal{L}_{proj}.
    \end{aligned}
\end{equation}

\subsection{SMPL Solver}
In the last stage, we transform the 3D skeleton key points obtained from the previous stage into the world coordinate system and use a temporal attention-based network to predict the parameters of the SMPL. The loss function of SMPL Solver is as follows:
\begin{equation}
    \begin{aligned}
    \mathcal{L}_{smpl} = &\lambda_4 \mathcal{L}_{mse}(\beta) + \lambda_5 \mathcal{L}_{mse}(\theta) \\&+ \lambda_6 \mathcal{L}_{mse}(J_{smpl}) + \lambda_7 \mathcal{L}_{mse}(V_{smpl}),
    \end{aligned}
\label{equ:loss}
\end{equation}
where $J_{smpl}$ and $V_{smpl}$ are generated by SMPL:
\begin{equation}
    \begin{aligned}
        J_{smpl}, V_{smpl} = \text{SMPL}(\beta, \theta, \tau).
    \end{aligned}
\end{equation}

Besides, we utilize an attention-based network for predicting the translation $T(n,t)$, within which we predict the distance $Tr(n,t)$ from the centroid of the point cloud $\overline{x}_{pc}(n, t)$ to the root node, mirroring the methodology employed in LiveHPS\cite{ren2024livehps}. The translation loss is defined as below:
\begin{equation}
    \begin{aligned}
        \mathcal{L}_{mse}(Tr) = &= \parallel \hat{Tr} - Tr^{GT}\parallel_2^2,
    \end{aligned}
\end{equation}
where $Tr^{GT} = T^{GT} - \overline{x}_{pc}^{GT}$.

\section{Experiment}
In this section, we compare our FreeCap with current LiDAR-based SOTA method LiveHPS~\cite{ren2024livehps} and camera-based SOTA method WHAM~\cite{wham:cvpr:2024} in Human-M3~\cite{fan2023human} and FreeMotion~\cite{ren2024livehps}. In this experiment, we utilize only the indoor portion of the FreeMotion dataset, which includes multi-view camera data. The extensive experiments demonstrates the effectiveness and robustness of our multi-modal fusion strategy. Furthermore, we also present comprehensive ablation studies to evaluate the necessity of our network modules and fusion method, and the efficiency and generalization of our matching strategy. Finally, we also discuss the universality of our freecap in various settings. Following LiveHPS, our evaluation metrics include J/V Err(PS/PST)($mm$), Ang Err($degree$), Accel Err($m/s^2$) and SUCD($mm$).
\begin{table*}[ht!]\small
\centering
\setlength\tabcolsep{2pt} 
\begin{tabular}{c|ccccc|ccccc}
\toprule
\multirow{2}{*}{} & \multicolumn{5}{c|}{Human-M3}& \multicolumn{5}{c}{FreeMotion}\\
\cmidrule(r){2-6}
\cmidrule(r){7-11}
& J/V Err(PS)$\downarrow$& J/V Err(PST)$\downarrow$&Ang Err$\downarrow$&Accel Err$\downarrow$&SUCD$\downarrow$ 
& J/V Err(PS)$\downarrow$& J/V Err(PST)$\downarrow$&Ang Err$\downarrow$&Accel Err$\downarrow$&SUCD$\downarrow$\\
\midrule
WHAM &69.42/83.35&-&10.13&9.20&-&82.83/97.55&-&12.24&4.51&-\\
WHAM(NV)&85.50/101.44&-&10.42&9.25&-&108.16/124.52&-&19.89&4.49&-\\
LiveHPS&57.81/71.27&97.11/103.10&10.44&12.58&6.75&59.30/73.12&100.81/109.09&13.10&6.18&4.97\\
Cali-based&58.18/71.24&98.37/104.76&10.06&9.69&6.70&57.33/69.37&99.36/106.24&12.53&5.99&5.00\\
\midrule
\textbf{Ours} &\textbf{55.45/68.52}&\textbf{ 96.47/102.67}&\textbf{9.14}&\textbf{9.60}&\textbf{6.66} &\textbf{53.31/65.50}&\textbf{95.97/102.91}&\textbf{11.14}&\textbf{5.97}&\textbf{4.82}\\
Ours(NV) &56.24/69.25&96.45/102.46&9.17&9.59&6.58 & 57.37/69.61 &99.20/106.15&11.74&6.08&4.96\\
\bottomrule
\end{tabular}
\caption{Comparison with state-of-the-art methods on various datasets. Notably, the ``Cali-based" method is our designed LiDAR-camera fusion method based on calibration information. ``NV" represents the novel view of the camera in the test dataset.}
\label{tab:compare}
\end{table*}

\subsection{Implementation Details}
We build our framework on PyTorch 2.0.0 and CUDA 11.8 and run the whole process on a server equipped with an Intel(R) Xeon(R) 444 E5-2678 CPU and 8 NVIDIA RTX3090 GPUs. For PCM, we set $\delta$ to 100, $\lambda_0$ to 0.1 and $n_{iter}$ to 2 to get a stable matching. During the training of SPO, we train the network over 500 epochs with batch size of 32 and sequence length of 32, using an initial learning rate of $10^{-4}$, and AdamW optimizer with weight decay of $10^{-4}$. We set $\lambda_1=1$, $\lambda_2=1$, $\lambda_3 = 0.01$ throughout our experiment. As for SMPL solver, we set $\lambda_4=1$, $\lambda_5=0.2$, $\lambda_6=10$ and $\lambda_7 = 1$. For the experiment, we consistently employed the SURREAL~\cite{varol17} dataset for pre-training throughout our experimental procedure, mirroring the approach adopted by LiveHPS. Following this, we pre-train the WHAM method on the SURREAL dataset and subsequently refine it through fine-tuning on both the Freemotion~\cite{ren2024livehps} and Human-M3\cite{fan2023human} datasets.
\subsection{Comparison}

\begin{table*}[ht]\small
\centering
\setlength\tabcolsep{2pt} 
\begin{tabular}{c|ccccc|ccccc}
\toprule
\multirow{2}{*}{} & \multicolumn{5}{c|}{Human-M3}& \multicolumn{5}{c}{FreeMotion}\\
\cmidrule(r){2-6}
\cmidrule(r){7-11}
& J/V Err(PS)$\downarrow$& J/V Err(PST)$\downarrow$&Ang Err$\downarrow$&Accel Err$\downarrow$&SUCD$\downarrow$ 
& J/V Err(PS)$\downarrow$& J/V Err(PST)$\downarrow$&Ang Err$\downarrow$&Accel Err$\downarrow$&SUCD$\downarrow$\\
\midrule
w/o MT-Refine &  59.01/ 73.07 & 103.44/ 110.03 & 9.29 & 9.94 & 8.00 &
- & - & - & - & - \\ 
w/o Temp-Int & 59.55/73.78 & 103.79/ 110.60 & 9.35 & 9.71 & 8.06 &
55.48/68.09 & 97.54/104.81 & 11.47 & 6.20 & 11.79 \\ 
w/o Sensor-Int & 58.33/72.16 & 98.38/ 105.22 & 9.96 & 9.56 & 6.66  &
59.84/73.40 & 100.87/108.90 & 12.91 & 6.08 & 11.74  \\
Body Pose &58.84/75.81&97.90/107.46&13.10&13.10&6.12& 56.97/70.48&98.43/106.64&12.65&6.55&5.08\\
Calibration &59.46/72.77&98.70/105.07&10.13&9.66&6.62&58.57/71.21&100.17/107.38&12.23&6.15&4.98\\
\midrule
\textbf{Ours} &\textbf{55.45/68.52}&\textbf{96.47/102.67}&\textbf{9.14}&\textbf{9.60}&\textbf{6.66}&\textbf{53.31/65.50}&\textbf{95.97/102.91}&\textbf{11.14}&\textbf{5.97}&\textbf{4.82}\\
\bottomrule
\end{tabular}
\caption{Ablation study for our network modules and multi-modal fusion methods.}
\label{tab:network_compare}
\end{table*}

\begin{figure}[ht!]
	\centering
	\includegraphics[width=\linewidth]{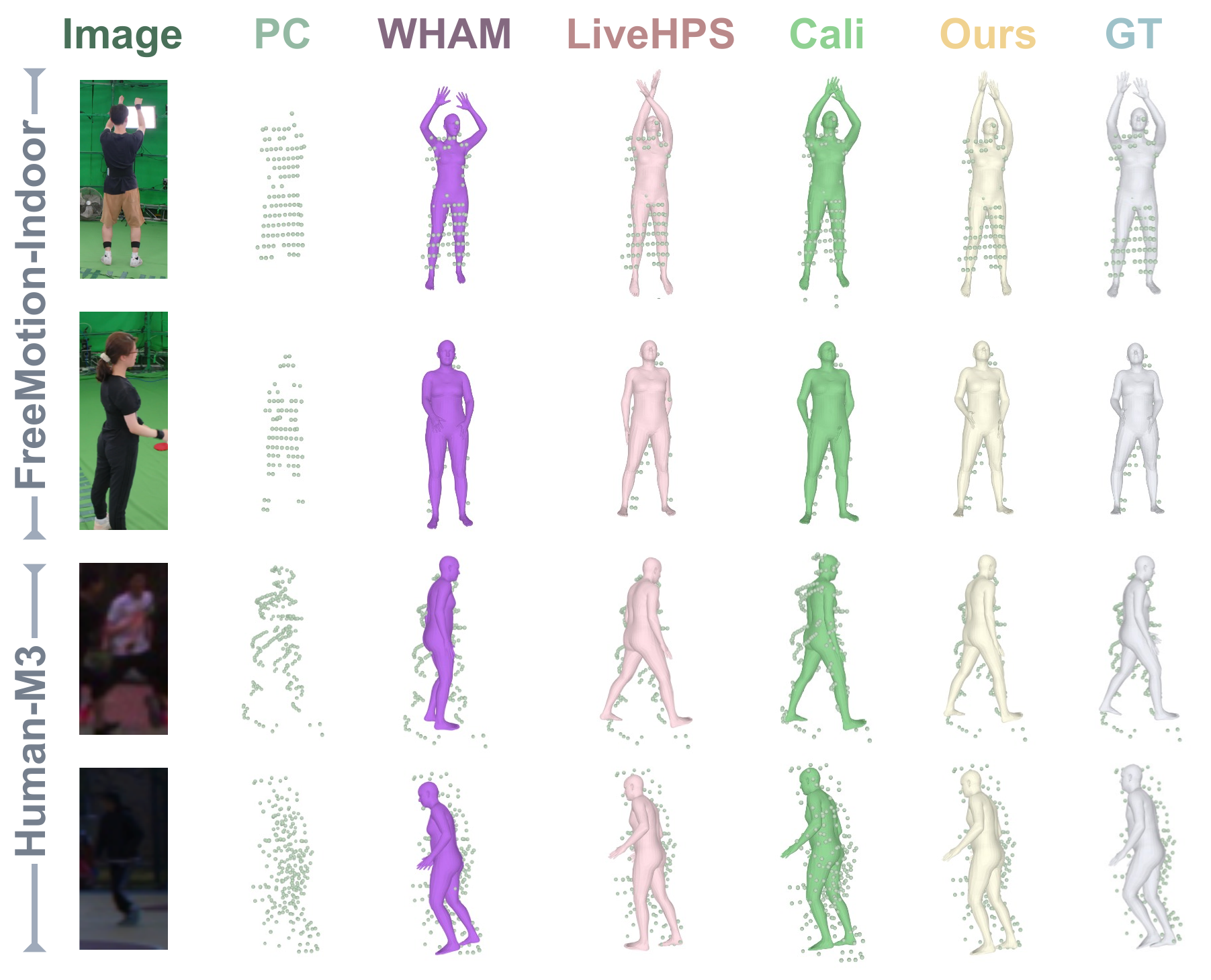}
	\caption{Qualitative comparisons. We show the global human mesh with point cloud, the point cloud matches the result better, representing more accurate estimation.}
	\label{fig:compare_vis}
\end{figure}

We evaluate our Freecap with SOTA method LiveHPS, WHAM and our designed calibration-based method in the testing dataset of FreeMotion and Human-M3 to demonstrate the effectiveness in multi-person large-scale scenes and the generalization in different novel views of cameras, here we only use single LiDAR and single camera as the input of FreeCap, the Human-M3 only provides the merged point cloud from all LiDARs. Our method achieves SOTA as shown in Tab.~\ref{tab:compare}, it is worth noting that WHAM estimates the translation based on the provided location in the first frame. It is unfair for other methods, which are information independent of the starting frame, so we do not show the WHAM's global evaluations. FreeCap achieves significant improvement, especially in J/V Err(PS) and Ang Err, which reflect the accuracy of human local pose, other global evaluations also improved based on the better local pose estimation. Since there is no calibration-based LiDAR-camera fusion method, we also design a simple calibration-based method based on the transformer framework, while FreeCap is efficient with more expandable sensor settings. To evaluate the generalization of our calibration-free strategy, we also evaluate WHAM and FreeCap in the testing dataset with the novel view of camera. FreeCap can maintain performance stability when input data from unknown perspectives, while the performance of WHAM deteriorates significantly. The visual comparisons presented in Fig.~\ref{fig:compare_vis} further underscore the superiority of our method in both pose and shape estimations, such as the first line, the LiveHPS estimates the incorrect human upper limb pose because of the occlusion in point cloud and WHAM estimates the incorrect human body shape because lack of depth information, while our fusion method utilizes the information from image to optimize the limb pose.

\subsection{Ablation Study}
We evaluate the superiority of each module in Human-M3 and FreeMotion. To prove the efficiency of our matching algorithm, we evaluate the matching accuracy and the frames per second(FPS) when inference in Human-M3. Finally, we also evaluate our sensor-expandable network framework in FreeMotion by adding more cameras.

\noindent\textbf{Network Architecture.}
As Tab.~\ref{tab:network_compare} shown, the PCM module without matching refinement results in incorrect human matching pairs because of the similar body pose in the scenes, incorrect matching results in a significant effect for our calibration-free multi-modal fusion. In our SPO, we utilize the temporal interaction and the multi-sensor interaction, the result proves the significant improvements in both interactions. The more direct fusion strategy is to use the body pose estimated from camera and LiDAR for fusion, since the body pose is perspective-independent intermediate results, but it lost the raw data information. Moreover, FreeCap can predict the calibration matrix for calibration-based fusion, but the method cannot dynamically learn the useful information in different sensors.

\begin{figure}[ht!]
	\centering
	\includegraphics[width=\linewidth]{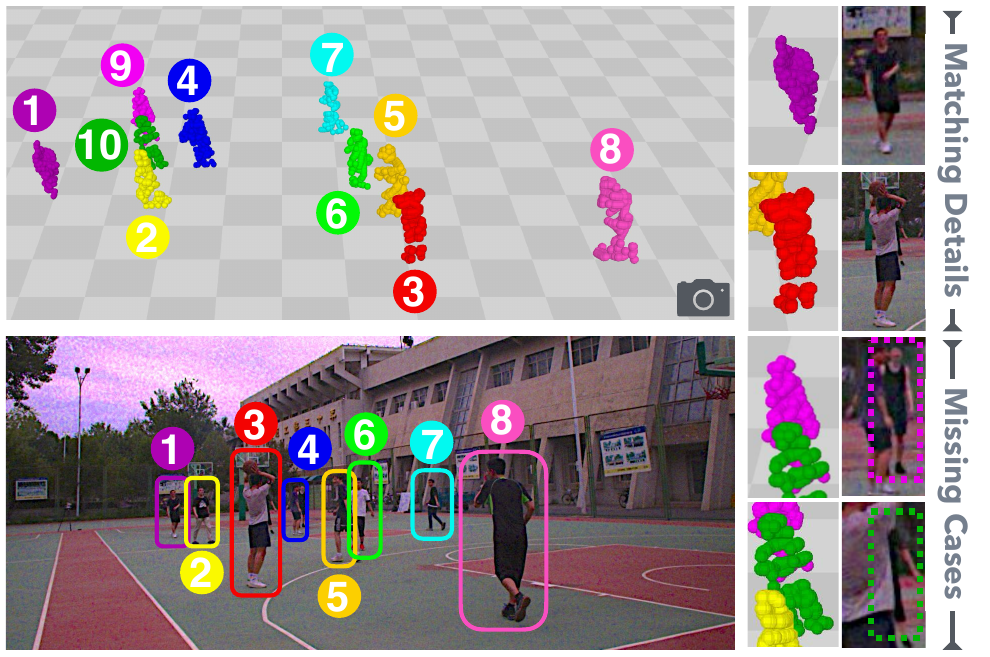}
	\caption{Visualization of our matching results in Human-M3. The view of camera and LiDAR is different, while the location of camera is labeled by the camera logo. We zoom in on some cases on the right.}
	\label{fig:compare_match_vis}
\end{figure}
\noindent\textbf{Matching Strategy.}
To further demonstrate the accuracy and efficiency of our matching method, we conduct a detailed ablation study on the PCM, as shown in Tab.~\ref{tab:match}. The ``KPs" represents we calculate all 2D key points and all 3D key points in every possible matching pair, which requires the time complex is $\bigO(n^n)$. The ``KP" represents we random select 3D key points of one person and calculate with all 2D key points, the time complex is $\bigO(n)$ but the accuracy decrease. ``Pose" represents we only use the estimated body pose for single frame matching and ``P \& T" represents we use the sequential body pose for sequence matching. ``P \& K" represents our matching method for single frame matching without temporal information. Our method achieves the highest accuracy while balancing efficiency. Fig.~\ref{fig:compare_match_vis} demonstrates our method's performance in extremely complex scenes with ten individuals, our method works well even if some persons in the camera are missing.

\begin{table}[t!]\small
\centering
\setlength\tabcolsep{2pt} 
\begin{tabular}{c|ccccc|c}
\toprule
& KPs& KP & Pose & P \& K & P \& T & \textbf{P \& T \& K}\\
\midrule
Acc.(\%)$\uparrow$ & 89.88&77.07&81.59&85.87&96.01&\textbf{98.10}\\
FPS$\uparrow$ & 0.08&12.06&127.47&3.40&309.28&\textbf{159.49}\\
\bottomrule
\end{tabular}
\caption{Ablation study for our matching method in Human-M3.}
\label{tab:match}
\end{table}

\section{Conclusion}
In this paper, we introduce a novel hybrid MoCap system that works in open environments without calibration, leveraging single LiDAR and expandable cameras. Our approach integrates the strengths of LiDAR and camera-based methods, utilizing cross-modal matching to effectively align and predict human poses in multi-person scenes. By combining 2D and 3D key points and optimizing the calibration matrix, our method adjusts matching conditions dynamically, ensuring accurate and robust motion capture. The extensive experiments demonstrate that our method achieves SOTA, offering a flexible and expandable solution for complex motion capture scenarios.

\section*{Acknowledgments}
This work was supported by NSFC (No.62206173), Shanghai Frontiers Science Center of Human-centered Artificial Intelligence (ShangHAI), MoE Key Laboratory of Intelligent Perception and Human-Machine Collaboration (KLIP-HuMaCo).
%

\bibliography{aaai25}

\end{document}